\documentclass[]{article}
\usepackage{proceed2e}
\usepackage{amsfonts}
\usepackage{amsmath}
\usepackage[authoryear]{natbib}
\usepackage{graphicx} 
\usepackage{tabularx}
\usepackage[table]{xcolor}
\usepackage{bm}

\title{An Improved Three-Weight Message-Passing Algorithm} 

\author{} 

\author{ {\bf Nate~Derbinsky} \\
Disney Research Boston \\  
222 3rd Street\\ 
Cambridge, MA 02142 \\ 
\And 
{\bf Jos\'{e}~Bento}  \\ 
Disney Research Boston          \\ 
222 3rd Street \\
Cambridge, MA 02142 \\              
\And 
{\bf Veit~Elser}  \\
Department of Physics \\          
Cornell University    \\    
Ithaca, NY 14850 \\
\And
{\bf Jonathan S.~Yedidia} \\
Disney Research Boston \\
222 3rd Street\\
Cambridge, MA 02142}       
 
\begin{document} 
 
\maketitle 
 
\begin{abstract} 
We describe how the powerful ``Divide and Concur'' algorithm for constraint
satisfaction can be derived as a special case of a message-passing version of 
the Alternating Direction Method of Multipliers (ADMM) algorithm for convex
optimization, and introduce an improved message-passing algorithm based on ADMM/DC
by introducing
three distinct weights for messages, 
with ``certain'' and ``no opinion'' weights, as well as the standard weight 
used in ADMM/DC. The ``certain'' messages allow our 
improved algorithm to implement constraint propagation as a special case, while
the ``no opinion'' messages speed convergence for some problems by making the algorithm focus
only on active constraints. We describe how our three-weight version of ADMM/DC
can give greatly improved performance for non-convex problems such as circle packing and 
solving large Sudoku puzzles, while
retaining the exact performance of ADMM for convex problems. We also describe the
advantages of our algorithm compared to other message-passing algorithms based
upon belief propagation.

\end{abstract} 
 
\section{INTRODUCTION} 
\label{sec:intro}

The Alternating Direction Method of Multipliers (ADMM) is a classic algorithm 
for convex optimization
dating from the 1970s \citep{Glowinski1975, Gabay1976}. The fact 
that it
is well-suited for {\em distributed} implementations was emphasized in a more recent 
extended review
by \citet{Boyd2011}. Perhaps owing to that review,
there has recently been an upsurge in interest in the ADMM algorithm, and 
it has been
used in applications including LP decoding of error-correcting codes 
\citep{Barman2011} and
compressed sensing \citep{Afonso2011}.

In this paper, we develop a message-passing algorithm based on ADMM suitable for 
{\em non-convex} as well as convex
problems (message-passing versions of ADMM for convex problems have already been developed by \citet{Martins2011}). At first sight, applying ADMM to non-convex problems
would seem to be a gross contradiction of the convexity assumptions underlying the derivation
of ADMM.
But in fact, ADMM turns out to be a well-defined algorithm for general optimization
functions, and, as we shall show, it is often a powerful heuristic algorithm even for NP-hard non-convex
problems. 

There is an analogy between ADMM and the famous belief propagation (BP)
 message-passing
algorithm in that they both can be used on problems far beyond those for which they give exact answers. 
Thus, while BP was originally derived and is 
only exact on cycle-free graphical models, it often turns out to
be a very powerful heuristic algorithm for graphical models with many cycles \citep{Yedidia2003}. 
Similarly, ADMM was originally
derived and is only guaranteed to be exact for convex objective functions, but is
 often a powerful
heuristic algorithm for a much wider variety of objective functions. Our message-passing version
of ADMM in fact has much in common with BP, but with some important advantages that we will describe below.

As we shall show, when the objective function being optimized by ADMM consists entirely of
``hard'' constraints, the ADMM algorithm reduces to the powerful ``Divide and Concur''
(DC) constraint satisfaction algorithm proposed by 
\citet{Gravel2008}. The DC algorithm has been shown to be effective
for a very wide variety of non-convex constraint satisfaction problems
\citep{Elser2007, Gravel2009} and has also been re-interpreted
as a message-passing algorithm \citep{YedidiaEtAl2011, Yedidia2011}. 
 
The message-passing version of ADMM that we develop here gives an intuitive explanation of 
how the algorithm functions---messages are sent back and forth between variables and
cost functions about the best values for the variables, until
consensus is reached. It also raises a natural question---how should those messages
be weighted?

We introduce here an extension of ADMM, that is well-suited to a variety of non-convex problems,
whereby the messages can have three different weights. A ``certain'' weight means that
a cost function or variable is certain about the value of a variable and all other 
opinions should defer to that value; a ``no opinion'' weight
means that the function or variable has no opinion about the value, 
while a ``standard weight'' means that
the message is weighted equally with all other standard-weight messages. 

The message-passing
version of the standard ADMM
algorithm will use only standard-weight messages; we will show that adding ``certain'' and
``no opinion'' weights gives a natural way to improve the algorithm for non-convex problems.
In particular, we show that for solving problems like Sudoku, ``certain'' weights can let
the algorithm implement constraint propagation \citep{Tack2009}, while for problems like
packing circles into a square, ``no opinion'' weights improve convergence by letting
the algorithm ignore ``slack'' 
constraints and focus on active ones.

Although our three-weight version of the ADMM algorithm only constitutes a relatively 
small change to the standard ADMM algorithm, it gives enormous
improvements in the rate at which the algorithm finds solutions for a variety of 
different problems. In this paper we have focused on the Soduku and circle packing
problems for purposes of illustration, but we emphasize that the three-weight 
algorithm can be applied to a wide range of other convex or non-convex
optimization problems. 

\section{MESSAGE-PASSING ADMM}
\label{sec:factorgraph}
 
We begin by deriving a message-passing version of the ADMM algorithm, and show that it can be applied
to a completely general optimization problem. Suppose that we are
given the general optimization problem of finding a configuration of some variables that minimizes 
some
objective function,
subject to some constraints. The variables may be continuous or discrete, but we will represent all variables
as continuous, which we can do by adding constraints that enforce the discrete nature
of some of the variables. For example, if we have a variable that can take on $Q$ discrete labels, we could
represent it with $Q$ binary indicator variables representing whether the original variable is in
one of the $Q$ states, and each binary variable is represented as a continuous variable 
but subject to the constraint
that exactly one of the $Q$ indicator variables
is equal to $1$ while the rest are equal to $0$.

We can collect all of our $N$ continuous variables, and represent them as a vector
$r \in \mathbb{R}^N$, and our problem becomes one of minimizing an objective function $E(r)$, subject
to some constraints on $r$. We can consider all the constraints to be part of the objective function by
introducing a cost function $E_a(r)$ for each constraint, such that $E_a(r) = 0$ if the constraint is 
satisfied, and $E_a(r) = \infty$ if the constraint is not satisfied. The original objective function
$E(r)$ may also be decomposable into a collection of local cost functions. For example, for the problem
of minimizing an Ising Model objective function, there would be a collection of 
local cost functions representing
the ``soft'' pairwise interactions and local fields, and there would be other cost functions representing
the the ``hard'' constraints that each Ising variable can only take on two possible values.

\begin{figure}[htbp]
\begin{center}
\includegraphics[width=.95\columnwidth]{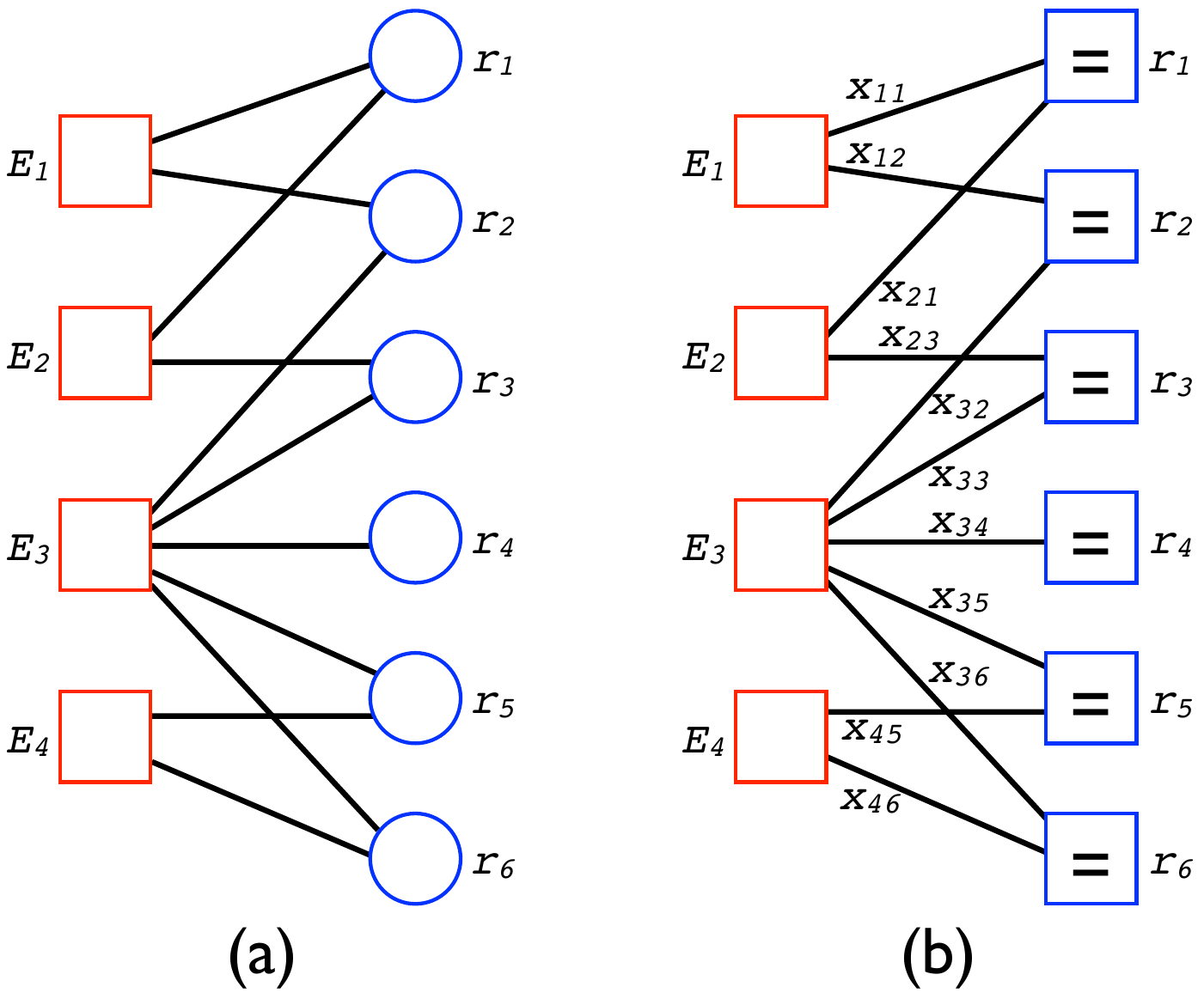}
\end{center}
\caption{Factor graph representations of an optimization problem. (a) The blue circles represent the variables, while
the red squares represent hard or soft cost functions. If a line connects a square to circle, that means
that the corresponding cost function depends on the corresponding variable. (b) A Forney factor graph representing
the same problem. Each of the original $r_j$ variable nodes is converted into an equality constraint, and
the $r_j$ variable is replaced by
copies denoted $x_{ij}$ that sit on the edges.}
\label{fig:factorgraph}
\end{figure}

To summarize, then, we focus on the general problem of minimizing an objective function written as
$ \sum_{a=1}^M E_a(r) $
where there are $M$ cost functions that can be either ``soft'' or ``hard'', and $r \in \mathbb{R}^n$.
Such a problem can be given a standard ``factor graph'' \citep{Kschischang2001, Loeliger2004} 
representation like that in Figure \ref{fig:factorgraph}(a).

To derive our message-passing algorithm, we will manipulate the problem into a series of equivalent forms, before
actually minimizing the objective. The first manipulation is to convert our problem over the variables $r_j$ into
an equivalent problem that depends on variables $x_{ij}$ that sit on the edges of a ``normalized'' Forney-style
factor graph \citep{Forney2001}, as in Figure \ref{fig:factorgraph}(b). The variables in the standard factor graph are replaced with
equality constraints, and each of the edge variables is a copy of the corresponding variable that was on its right.
The point is that edge variables attached to the same equality constraint must ultimately equal each other, but
they can temporarily be unequal while they separately try to satisfy different cost functions on the left. We will
use the notation $x$ to represent a vector consisting of the entire collection of $x_{ij}$ edge variables; note that
$x$ normally has higher dimensionality than $r$.

\begin{figure}[htbp]
\begin{center}
\includegraphics[width=.95\columnwidth]{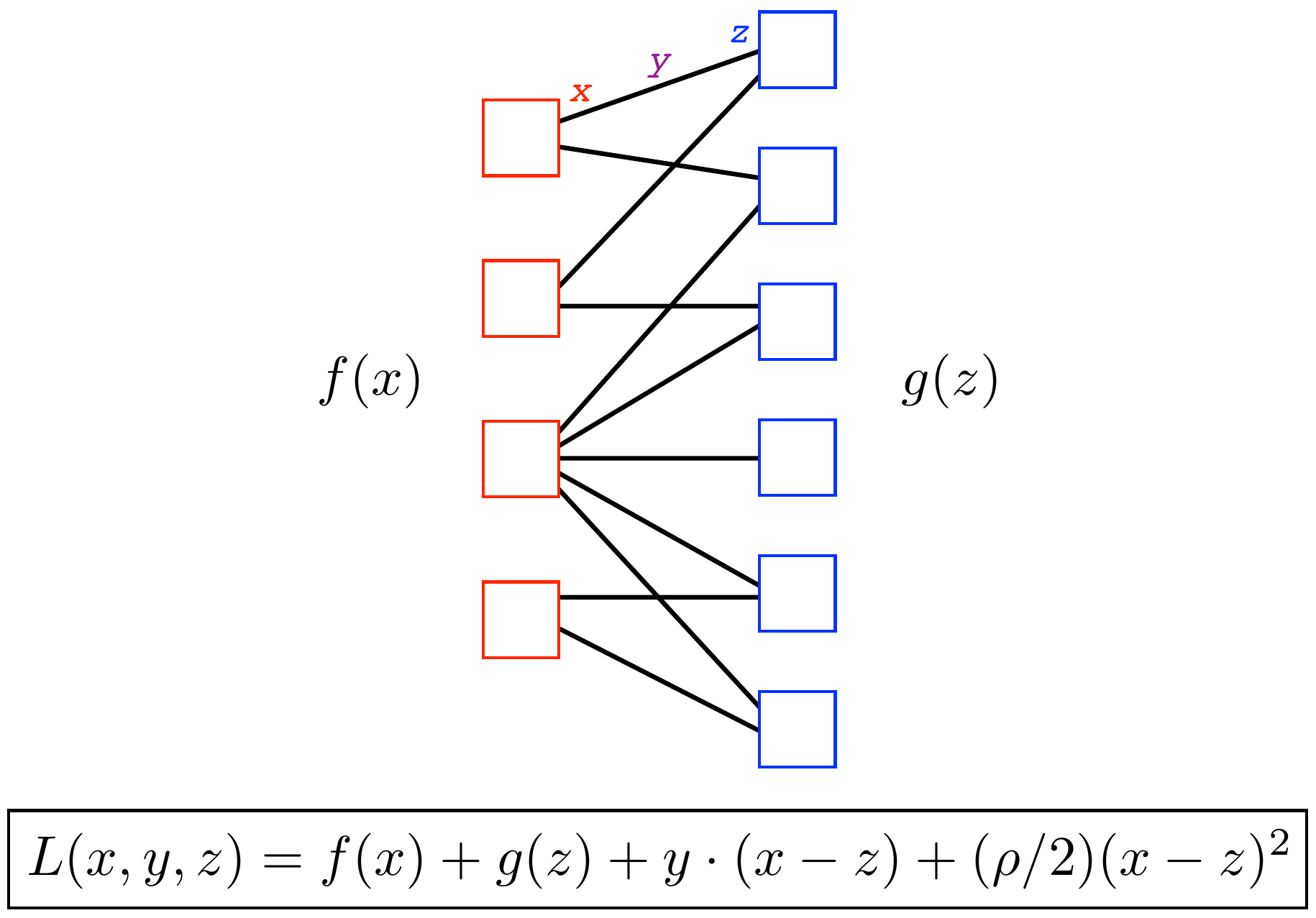}
\end{center}
\caption{Solving constrained optimization problems is equivalent to a minimization problem
that naturally splits into three pieces: (1) minimizing the original soft and hard cost functions $f(x)$ on the
left, (2) minimizing the equality cost functions $g(z)$ on the right, and (3) ensuring that the $x = z$ with the
Lagrange multipliers $y$.}
\label{fig:bipartite}
\end{figure}

Because of the bipartite structure of a Forney factor graph, 
we can split our cost functions into two groups: those on the
left that represent our original soft cost functions and hard constraints, and those on the right that represent
equality constraints. We now imagine that each $x_{ij}$ edge variable sits on the left side of the edge,
and make a copy of it called $z_{ij}$ that sits on the right side of the edge, and formally split our
objective function $E(x)$ into a sum of the left cost functions $f(x)$ and the right cost functions $g(z)$, where
$z$ is a vector made from all the $z_{ij}$ variables (see Figure \ref{fig:bipartite}). 

The constraint
that each edge variable $x_{ij}$ equals its copy $z_{ij}$ will be enforced by a Lagrange multiplier $y_{ij}$, in 
a Lagrangian that we can write as $L(x,y,z) = f(x) + g(z) + y \cdot (x-z)$.  
It will turn out to be useful to add another term $(\rho/2) (x-z)^2$ to ``augment'' the Lagrangian. Since $x = z$ at the
optimum, this term is zero at the optimum and non-negative elsewhere, so it clearly does not change
the optimum. The parameter $\rho$ can be thought of as a scalar, but later we will generalize it
to be a vector with a different $\rho_{ij}$ for each edge. 
In summary, as illustrated in Figure \ref{fig:bipartite}, our original problem of minimizing $E(r)$ has become equivalent to finding the
minimum of the augmented Lagrangian 
\begin{equation}
L(x,y,z) = f(x) + g(z) + y \cdot (x-z) + \frac{\rho}{2} (x-z)^2 .
\end{equation}

To make progress in the derivation of our algorithm, 
we now make the assumption that each of the local cost functions $f_a(x)$ on the left
side of the factor graph is convex. We emphasize again
that our final algorithm will be well-defined even when this assumption is violated. All the equality
cost functions on the right are clearly convex, and the augmenting quadratic terms are convex, so our overall function
is convex as well \citep{Boyd2004}. That means that we can find the minimimum of our Lagrangian
by maximizing the dual function 
\begin{equation}
h(y) = L(x^*, y, z^*)
\end{equation}
where $(x^*, z^*)$ are the values of $x$ and $z$ that minimize $L$ for a particular choice of $y$:
\begin{equation}
(x^*, z^*) = \underset{x,z}{\rm {argmin}} \, L(x,y,z)
\end{equation}

We use a gradient ascent algorithm to maximize $h(y)$. Thus, given values of $y^t$ at some iteration $t$,
we iteratively 
compute $(x^*,z^*) = \underset{x,z}{\rm {argmin}} \, L(x,y^t,z)$, and then move in the direction
of the gradient of $h(y)$ according to
\begin{equation}
y^{t+1} = y^t + \alpha \frac{\partial h}{\partial y} = y^t + \alpha ( x^* - z^* )
\end{equation}
where $\alpha$ is a step-size parameter. 

We take advantage of the bipartite structure of our factor graph to decouple the minimizations of $x$
and $z$. Introducing the scaled Lagrange multiplier $u = y / \rho$ \citep[see][sec. 3.1.1]{Boyd2011}
and the ``messages'' 
$m^t = x^t + u^t$ and $n^t = z^t - u^t$,
 we obtain the following iterative equations which define our message-passing version
of ADMM:
\begin{equation}
\label{eq:x_update}
x^{t} = \underset{x} {\rm{argmin}} \, \left[ f(x) + (\rho/2) (x-n^t)^2 \right]
\end{equation}
\begin{equation}
\label{eq:z_update}
z^{t+1} = \underset{z} {\rm{argmin}} \, \left[ g(z) + (\rho/2) (z-m^t)^2 \right]
\end{equation}
\begin{equation}
\label{eq:u_update}
u^{t+1} = u^t + \frac{\alpha}{\rho} ( x^t - z^{t+1} )
\end{equation}

The algorithm is initialized by choosing $u^0 = 0$, and starting with some initial $z^0$. Then equation 
(\ref{eq:x_update}) determines $x^0$, equation (\ref{eq:z_update}) determines $z^1$, equation (\ref{eq:u_update})
determines $u^1$, we go back to equation (\ref{eq:x_update}) for $x^1$, and so on.

Intuitively, the $z$ and $x$ variables are analogous to the single-node and multi-node ``beliefs'' in belief propagation, while the messages $m$ and $n$ are messages from the nodes on the left to those on the right, and vice-versa, respectively, much as in the ``two-way'' version of the standard belief propagation algorithm \citep{Yedidia2005}. 
In each iteration, the algorithm computes beliefs on the left based on the messages coming from the right, then beliefs on the right based on the messages from the left, and then tries to equalize the beliefs on the left and right using the $u$ variables. 
Notice that the $u$ variables keep a running total of the differences between the beliefs on the left and the right, and are much like control variables tracking a time-integrated difference from a target value.

It is very important to realize that all the updates in these equations are {\em local} computations and
can be done in parallel. Thus, if function cost $a$ is connected to a small set of edges with variables $\{x\}_a$, then 
it will only
need to look at the messages $\{n\}_a^t$ on those same edges to perform its local computation
\begin{equation}
\label{eq:local_x_update}
\{x\}_a^{t+1} = 
\underset{\{x\}_a}{\rm{argmin}}
\left[ f_a(\{x\}_a)
+ (\rho / 2)
(\{x\}_a
- \{n\}_a^{t}
)^2 \right].
\end{equation}

These local computations are usually easy to implement with small ``minimizing'' 
subroutines specialized to the particular $f_a(\{x\}_a)$.
Such a subroutine balances the desire to minimize the local $f_a(\{x\}_a)$ with the desire to
agree with the $\{n\}_a^t$ messages coming from other nodes. The $\rho$ parameter lets us
vary the relative strength of these competing influences.

The minimizations in equation~(\ref{eq:z_update}) are similarly all local computations that can be done in parallel. In fact,
because the $g_a(\{z\}_a)$ functions on the right all represent 
hard equality constraints, these minimizations will reduce to a particularly simple form (the output $z$'s will be
given as the mean of the incoming $m$ messages), as we shall describe
in the next section.

When all the $f_a(\{x\}_a)$ functions are convex, our overall problem is convex, and the ADMM algorithm
provably converges to the correct global minimum \citep{Boyd2011}, although it is important to note that no guarantees are
made about the speed of convergence.\footnote{Our derivation starting with an arbitrary optimization problem and using
Forney factor graphs guarantees that the $g_a(\{z\}_a)$ functions on the right are
all equality constraints, which are convex. More generally, ADMM can be considered to be an algorithm which operates on any
functions $f(x)$ and $g(z)$, and gives exact answers so long as $f(x)$ and $g(z)$ are both convex, as in \citep{Boyd2011}.}  
 However, as is easy to see, the algorithm is in fact perfectly well
defined even for problems where the $f_a(\{x\}_a)$ functions are not convex (so long as they are bounded below).
In the next section, we begin our investigation of how the algorithm might be used for non-convex problems, beginning with its
relation to the Divide and Concur algorithm.

\section{DIVIDE AND CONCUR}
\label{sec:dc}

The Divide and Concur (DC) algorithm \citep{Gravel2008} for constraint satisfaction problems
historically traces its roots back to the 
Douglas-Rachford algorithm \citep{Douglas1956}, later extended into a projection operator splitting method for solving
convex problems by \citet{Lions1979}. The surprisingly successful application
of such a ``difference-map'' projection algorithm to the {\it non-convex} phase retrieval problem by
\citet{Fienup1982} led to the realization that these algorithms could also be useful for non-convex problems
\citep{Bauschke2002, Elser2007}.

Our object in this section is to make clear that the DC algorithm is in fact a special case of the ADMM algorithm, 
for those problems where
the $f_a(\{x\}_a)$ all represent hard constraints. In that case, we can write the legal configurations of $\{x\}_a$ as
a constraint set $\mathcal{D}$, and require that $f_a(\{x\}_a)=0$ for $\{x\}_a \in \mathcal{D}$, and 
$f_a(\{x\}_a)=\infty$ for $\{x\}_a \notin \mathcal{D}$.
Then each local minimization to compute $\{x\}_a^{t+1}$ as given by equation (\ref{eq:local_x_update}) would reduce to a projection of the incoming messages $\{n\}_a$ onto the constraint set:
\begin{equation}
\{x\}_a^{t+1} = P_D(\{n\}_a^t).
\end{equation}
This is easily understood if one realizes that the $f_a(\{x\}_a)$ term in equation (\ref{eq:local_x_update}) enforces that $\{x\}_a$ must be in the constraint set, while minimizing the $(\rho/2) (\{x\}_a - \{n\}^t_a)^2$ term enforces that the computed $\{x\}_a$ values are as close as possible (using a Euclidean metric) to the messages $\{n\}_a^t$, and that is the definition of a projection.

Similarly the cost functions $g_a(\{z\})$ represent hard equality constraints, so we can write the $\{z\}_a^{t+1}$ updates as projections of the $\{m\}_a^t$ messages onto the equality constraint sets $\mathcal{C}$:
\begin{equation}
\{z\}_a^{t+1} = P_C(\{m\}_a^t).
\end{equation}
Assuming that all the $\rho$ weights are equal (we will go beyond this assumption in the next section), this can be further simplified: the 
$\{z\}_a^{t+1}$ values on the edges connected to an equality constraint node should all equal the mean of the messages incoming to that equality node. 

Now if we choose the step-size parameter $\alpha$ to equal the weight $\rho$, we can further simplify, and show that instead of requiring updates of all the variables $x^t$, $z^t$, $u^t$, $m^t$, and $n^t$, our algorithm actually reduces to an iteration of a single state variable: the messages $m^t$. 
Some tedious but straightforward algebra manipulating equations (\ref{eq:x_update}), (\ref{eq:z_update}), and (\ref{eq:u_update}), along with the definitions of $m^t$ and $n^t$, lets us eliminate the $x^t$, $z^t$, and $u^t$ variables and reduce to the message-update equations:
\begin{equation}
n^{t+1} = 2 P_C(m^t) - m^t
\end{equation}
and 
\begin{equation}
m^{t+1} = m^t + P_D(n^{t+1}) - P_C(m^t)
\end{equation}
or equivalently leave us with the single update equation for the $m^t$ messages:
\begin{equation}
\label{eq:difference_map}
m^{t+1} = P_D(2P_C(m^t) - m^t) - (P_c(m^t) - m^t)
\end{equation}
Equation (\ref{eq:difference_map}) is also known as the ``difference-map'' iteration used by the Divide and Concur algorithm.\footnote{Sometimes Divide and Concur uses the dual version of the difference-map, where $n^t$ is updated according to a rule obtained
from Equation ({\ref{eq:difference_map}}) by swapping the $P_C$ and $P_D$ projections.}

\section{THREE-WEIGHT ALGORITHM}
\label{sec:threeweight}

Notice that although the DC algorithm is a special case of the ADMM algorithm, the weights $\rho$ in ADMM have disappeared in the DC update equations.
These weights have a very intuitive meaning in our message-passing ADMM algorithm---they reflect how strongly the messages should be adhered to in comparison to the local function costs; i.e. the ``reliability'' or ``certainty'' of a message.
It is thus natural to consider a generalized version of the message-passing ADMM algorithm where each edge $(ij)$ connecting a function cost $i$ to an equality node $j$ is given its own value of $\rho_{ij}$ reflecting the certainty of messages on that edge. 

In fact, it is more natural to consider an even greater generalization, with different weights for messages going to the left and those going to the right, and where the weights can change with each iteration. 
We denote the vector of weights going to the left at time $t$ as $\overleftarrow{\rho}^t$, and similarly the
vector of weights going to the right are denoted $\overrightarrow{\rho}^t$. 
When we want to denote a particular weight on an edge $(ij)$, we will denote it $\overleftarrow{\rho}_{ij}^t$ or $\overrightarrow{\rho}_{ij}^t$.
We need to be careful that for a convex problem, we ensure that leftward and rightward weights eventually equal each other and are constant, 
because otherwise such an algorithm will not necessarily converge to the global optimum.\footnote{We observed empirically that variants of our algorithm would fail for convex problems when the weights in the two directions were not eventually equal
or when they were not eventually constant. 
Also, the convergence proof for ADMM on convex problems presented in \citep{Boyd2011} can be generalized straightforwardly to different weights on each edge, but it depends on the weights being constant within an iteration and between iterations.}

We therefore need to be relatively conservative in modifying the algorithm, 
and present here a relatively simple modification which allows for only three possible values for the weights on each edge. 
First we have standard weight messages with some weight $\rho_0$ that is greater than zero and less than infinity.
The exact value of $\rho_0$ will be important for the rate of convergence for problems with soft cost functions, but it will be irrelevant for problems consisting entirely of hard constraints, just as it is in standard DC, so for simplicity one can suppose that $\rho_0=1$ for those problems. 
Second we allow for infinite-weight messages, which intuitively represent that the message's value is certainly correct.
Finally we allow for zero-weight messages, which intuitively represent that a function cost node or equality node is completely uncertain about the value that a variable should have, and its opinion should be ignored. 

In modifying the ADMM message-passing algorithm to allow for zero weights or infinite weights, we need to also be careful to properly deal with updates of $u_{ij}$ variables. 
Intuitively, $u_{ij}$ variables are tracking the ``disagreement'' between the left and right beliefs on an edge $(ij)$.
The $u_{ij}$ variable on an edge will grow in magnitude over time if the left belief $x_{ij}$ is persistently less than or persistently greater than the right belief $z_{ij}$. 
Because the $u_{ij}$ variables are added or subtracted to the beliefs to form the messages, the message values can become quite different from the actual belief values, as the $u_{ij}$ variables try to resolve the disagreement.
With infinite and zero weight messages, it is important to be able to ``reset'' the $u_{ij}$ variables to zero if there is no disagreement on that edge; for example when a infinite weight message is sent on an edge, it means that the message is certainly correct, so any previous disagreement recorded in the $u_{ij}$ should be ignored.

\subsection{DETAILED DESCRIPTION OF THREE-WEIGHT ALGORITHM}
For clarity, rather 
than provide a pseudo-code description of the algorithm, we provide a fully explicit English language
description.

We initialize by setting all $u^0_{ij}= 0$, and normally use zero weights (e.g. $\overleftarrow{\rho}^0_{ij} = 0$ on each edge) for the initial $n^0$ messages from the right to left for those variables we
have no information about.
For any variables about which we are certain, we would accompany their messages with infinite weights.
We next compute $x^0$ using the standard update equations
\begin{equation}
\label{eq:x_updatewithrho}
x^{t} = \underset{x} {\rm{argmin}} \, \left[ f(x) + ({\overleftarrow{\rho}}^t/2) (x-n^t)^2 \right].
\end{equation}
Any ties in the $x^t$ updates are broken randomly. 
The left-to-right $m^t$ messages  are computed using $m^t = x^t + u^t$, but since $u^0=0$, the initial messages to the right will equal the initial $x^0$ beliefs.

The outgoing ${\overrightarrow{\rho}}^t$ weights are computed using an appropriate logic for the function cost on the left, which will depend on an analysis of the function. 
For example, for the Sudoku problem (see next section) we will only send out standard weights or infinite weights, depending on a logic that sends out infinite weights only when we are certain about the
corresponding $x$ value.
Whenever an infinite weight is used, whether for a right-going message at this point 
or for a left-going message at another stage, 
the $u_{ij}^t$ for that edge is immediately re-set to zero.

We next compute the $z^{t+1}$ right beliefs by taking a weighted average of the $m^t$ messages, weighted by the ${\overrightarrow{\rho}}^t$ weights. 
That means that if any message has infinite weight, it will control the average, and any zero-weight message will not contribute to the average.
If the logic used to send infinite weights is correct, there cannot be any contradictions between infinite weight messages.

To compute the weights coming back to the left from an equality node, we follow the following logic. 
First, if any edge is sending in an infinite $\overrightarrow{\rho}_{ij}^t$ weight, all edges out of the equality node get back an infinite $\overleftarrow{\rho}_{ij}^{t+1}$ weight.
Otherwise, all edges get back a standard $\overleftarrow{\rho}_{ij}^{t+1}$ weight as long as at least one of the incoming weights is non-zero. 
Finally, if all incoming $\overrightarrow{\rho}_{ij}^t$ are zero, the outgoing $\overleftarrow{\rho}_{ij}^{t+1}$ weights are also set to zero.

Next, all $u$ variables are updated. 
Any $u_{ij}$ on an edge that has an infinite weight in either direction is reset to zero. 
Also any edge that has a zero weight $\overrightarrow{\rho}^t_{ij}$ has its $u$ variable 
reset to zero (the reasoning is that it did not contribute to the average, and should agree with the consensus of the rest of the system). 
Any edge that has a standard weight $\overrightarrow{\rho}^t_{ij}$ while all other edges into its equality node have zero weight also has its $u$ variable reset to zero (the reasoning again is that there
was no disagreement, so there is no longer any need to modify the right belief). 
Any other edge that has a standard weight $\overrightarrow{\rho}^t_{ij}$  and a standard
weight $\overleftarrow{\rho}^{t+1}_{ij}$ will have its $u^{t+1}_{ij}$ updated according to the formula
$u^{t+1}_{ij} = u^t_{ij} +  (\alpha / \rho_0) ( x^t_{ij} - z^{t+1}_{ij})$. 
Once the $u$ variables are updated, we can update all the right-to-left $n^{t+1}$ messages according to the formulas $n^{t+1}_{ij} = z^{t+1}_{ij} - u^{t+1}_{ij}$.

Finally, we are done with an iteration and can go on to the next one. Our stopping criterion is that all the $n$ and $m$ messages are identical from iteration to iteration, to some specified numerical tolerance.

We now illustrate the utility of non-standard weights on two non-convex problems: Sudoku and circle-packing. 
We will find that infinite weights are useful for Sudoku, because they allow the algorithm to propagate certain information, while zero weights are useful for circle-packing because they allow the algorithm to ignore irrelevant constraints.

\section{SUDOKU}
\label{sec:sudoku}

A Sudoku puzzle is a partially completed row-column grid of cells partitioned into $N$ regions, each of size $N$ cells, to be filled in using a prescribed set of $N$ distinct symbols, such that each row, column, and region contains exactly one of each element of the set.
A well-formed Sudoku puzzle has exactly one solution.
Sudoku is an example of an exact-cover constraint-satisfaction problem and is NP-complete when generalized to ${N}\times{N}$ grids \citep{Yato2003}.

\begin{figure}[tb]
\begin{center}
\includegraphics[width=.5\columnwidth]{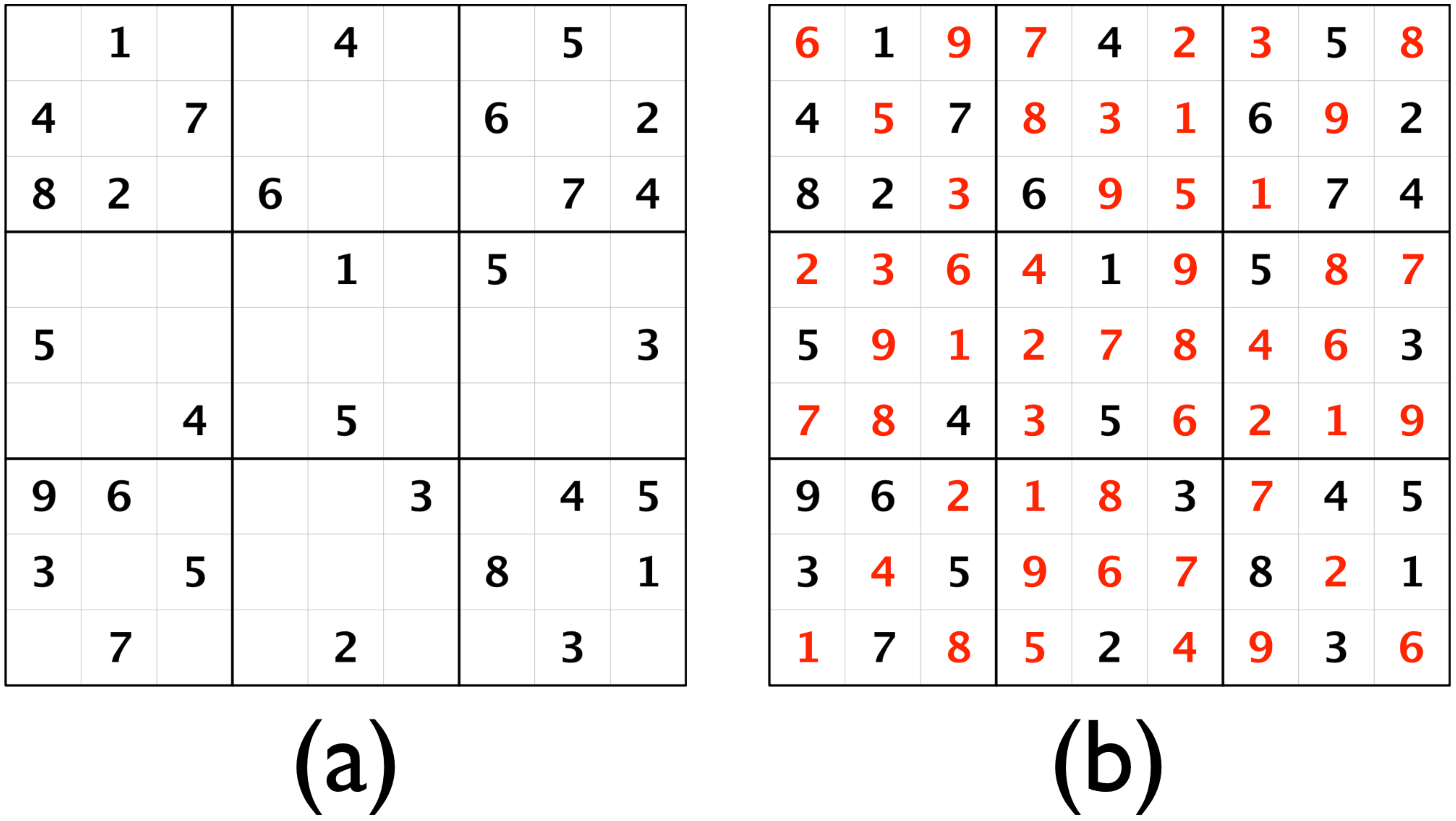}
\end{center}
\caption{A typical ${9}\times{9}$ sudoku puzzle: (a) original problem and (b) corresponding solution, with added digits marked in red.}
\label{fig:sudoku}
\end{figure}

People typically solve Sudoku puzzles on a ${9}\times{9}$ grid (e.g. see Figure \ref{fig:sudoku}) containing nine ${3}\times{3}$ regions, but larger square-in-square puzzles are also possible when investigating Sudoku-solving algorithms.
To represent an ${N}\times{N}$ square-in-square Sudoku puzzle as an optimization problem we use $\mathcal{O}(N^3)$ binary indicator variables and $\mathcal{O}(N^2)$ hard constraints.
For all open cells (those that have not been supplied as ``clues''), we use a binary indicator variable, designated as $v({row}, {column}, {digit})$, to represent each possible digit assignment. 
For example, the variables $v(1, 3, 1)$, $v(1, 3, 2)$ \ldots $v(1, 3, 9)$ represent that the cell in row 1, column 3 can take values 1 through 9.
We then apply hard ``one-on'' constraints to enforce digit distinctiveness: a \emph{one-on} constraint requires that a single variable is ``on'' (1.0) and any remaining are ``off'' (0.0).
We apply \emph{one-on} constraints to four classes of variable sets: 

\begin{enumerate}
\item $\forall{r}\forall{c}$ \{$v({row}, {col}, {dig})$ : ${row}=r$, ${col}=c$\} \linebreak {one digit assignment per cell}
\item $\forall{r}\forall{d}$ \{$v({row}, {col}, {dig})$ : ${row}=r$, ${dig}=d$\} \linebreak {one of each digit assigned per row}
\item $\forall{c}\forall{d}$ \{$v({row}, {col}, {dig})$ : ${col}=c$, ${dig}=d$\} \linebreak {one of each digit assigned per column}
\item $\forall{s}\forall{d}$ \{$v({row}, {col}, {dig})$ : ${sq}({row},{col})=s$, ${dig}=d$\} \linebreak {one of each digit assigned per square}
\end{enumerate}

Prior work on formulating Sudoku puzzles as constraint-satisfaction problems \citep[e.g.][]{Simonis2005} has utilized additional, redundant constraints to strengthen deduction by combining several of the original constraints, but we only utilize this base constraint set.

Analysis of Sudoku as a dynamical system has shown that puzzle difficulty depends not only on the global properties of variable size and constraint density, but also positioning patterns of the clues.
Algorithmic search through solution space can be chaotic, with search times varying by orders of magnitude across degrees of difficulty \citep{Ercsey2012}.

Though Sudoku is not convex, we demonstrate in this section that ADMM is often an effective algorithm: it only converges to actual solutions, it often completes puzzles quickly, and it scales to large puzzle sizes.

We also integrate an implementation of infinite weights within the \emph{one-on} minimizers, which serves to reduce the search space of the problem instance.
We show that introducing certainty via our three-weight algorithm often improves time-to-solution, especially in puzzles where constraint propagation is sufficient to logically deduce most or all of the puzzle solution with no search required \citep{Simonis2005}. 
This is of course natural and to be expected---if we can reduce the effective size of the puzzle to be solved by first successively inferring certain values for cells (much as humans do when solving Sudoku puzzles), only a smaller difficult core will need to be solved using the standard weight messages. 

\subsection{EVALUATION}
We implemented each hard \emph{one-on} constraint as a cost function on the left.
For this class of constraint, minimizing equation \eqref{eq:local_x_update} involves a linear scan: select the sole ``on'' edge as that which is certain and ``on'' ($\overleftarrow{\rho}_{ij}^{t}=\infty$ and $n_{ij}^{t}=1.0$) or, in absence of such an edge, that with the greatest incoming message value and a standard weight.

Outgoing weights ($\overrightarrow{\rho}_{ij}^t$) default to $\rho_0$, with three exceptions.
First, if a single edge is certain and ``on'' ($\overleftarrow{\rho}_{ij}^{t}=\infty$ and $n_{ij}^{t}=1.0$), all outgoing assignments are certain.
Second, if all but a single incoming edge is certain and ``off'' ($\overleftarrow{\rho}_{ij}^{t}=\infty$ and $n_{ij}^{t}=0.0$), all outgoing assignments are certain.
Finally, incoming certainty for an edge is maintained in its outgoing weight ($\overleftarrow{\rho}_{ij}^{t}=\infty \Rightarrow \overrightarrow{\rho}_{ij}^{t}=\infty$).

We downloaded 185 ${N}\times{N}$ Sudoku puzzles where ${N}\in\{9, 16, 25, 36, 49, 64, 81\}$ from an online puzzle repository \footnote{http://www.menneske.no/sudoku/eng}.
For each puzzle instance we applied both the ADMM message-passing algorithm and our three-weight algorithm using five random seeds for initial conditions.

For all puzzle trials, both algorithms converged to the correct solution, but Table \ref{fig:sudoku-results} provides evidence that our three-weight algorithm improved performance.
The ``\% Improved $>2\times$'' column indicates the percentage of puzzle trials within each puzzle size for which our algorithm converged in fewer than 50\% as many iterations given the same initial conditions: by this definition, our algorithm improved more than $80\%$ of all trials as compared to ADMM.
The ``Median Speedup'' column refers to the improvement in iterations-to-solution for each trial: overall the median improvement for our algorithm was a $4.12\times$ reduction in iterations, with a maximum improvement of $61.21\times$ on a single puzzle.
While a total of 55 trials required more iterations to solve using our algorithm, when aggregated by puzzle size and difficulty (as labeled by the puzzle author), only two classes of puzzle suffered reduced performance: (a) two ``impossible'' ${16}\times{16}$ puzzles and (b) the hardest ${49}\times{49}$ puzzle.
It is likely with these difficulty ratings, constraint propagation was of little assistance, and thus both algorithms relied upon equivalent search methods within a chaotic space, but from different starting points.

\begin{table}[tb]
\caption{Algorithmic comparison of iterations-to-convergence in $N$x$N$ square-in-square Sudoku puzzles. Each puzzle was solved using 5 different random seeds and improvement is comparing our three-weight algorithm to ADMM with a single standard weight of 1.0. The value of ``speedup'' is computed as $({iterations}_\text{ADMM} /{iterations}_\text{three weight})$ and ``\% improved'' refers to the percentage of trials that reduced iterations-to-solution by more than $2\times$. \\} 
\label{fig:sudoku-results}
\begin{tabular}{ | m{0.8cm} | m{1.4cm} | m{2.25cm} | m{2.0cm} | m{0.1cm} }

\centering \cellcolor{black} \textbf{\textcolor{white}{\bm{$N$}}} & \centering \cellcolor{black} \textbf{\textcolor{white}{\# Puzzles}} & \centering \cellcolor{black} \textbf{\textcolor{white}{\% Improved $>2\times$}} & \centering \cellcolor{black} \textbf{\textcolor{white}{Median Speedup}} & \\

\centering \textcolor{white}{0}9 & \centering 50 & \centering \textcolor{white}{0}83.20\% & \centering $3.35\times$ & \\ \cline{1-4}
\centering 16 & \centering 50 & \centering \textcolor{white}{0}74.40\% & \centering $3.65\times$ & \\ \cline{1-4}
\centering 25 & \centering 50 & \centering \textcolor{white}{0}82.80\% & \centering $5.58\times$ & \\ \cline{1-4}
\centering 36 & \centering 25 & \centering \textcolor{white}{0}77.60\% & \centering $5.35\times$ & \\ \cline{1-4}
\centering 49 & \centering \textcolor{white}{0}5 & \centering \textcolor{white}{0}80.00\% & \centering $4.44\times$ & \\ \cline{1-4}
\centering 64 & \centering \textcolor{white}{0}4 & \centering 100.00\% & \centering $6.01\times$ & \\ \cline{1-4}
\centering 81 & \centering \textcolor{white}{0}1 & \centering \textcolor{white}{0}60.00\% & \centering $2.03\times$ & \\ \cline{1-4}

\end{tabular}
\end{table}

\section{CIRCLE PACKING}
\label{sec:circlepacking}

\begin{figure}[tb]
\begin{center}
\includegraphics[width=.5\columnwidth]{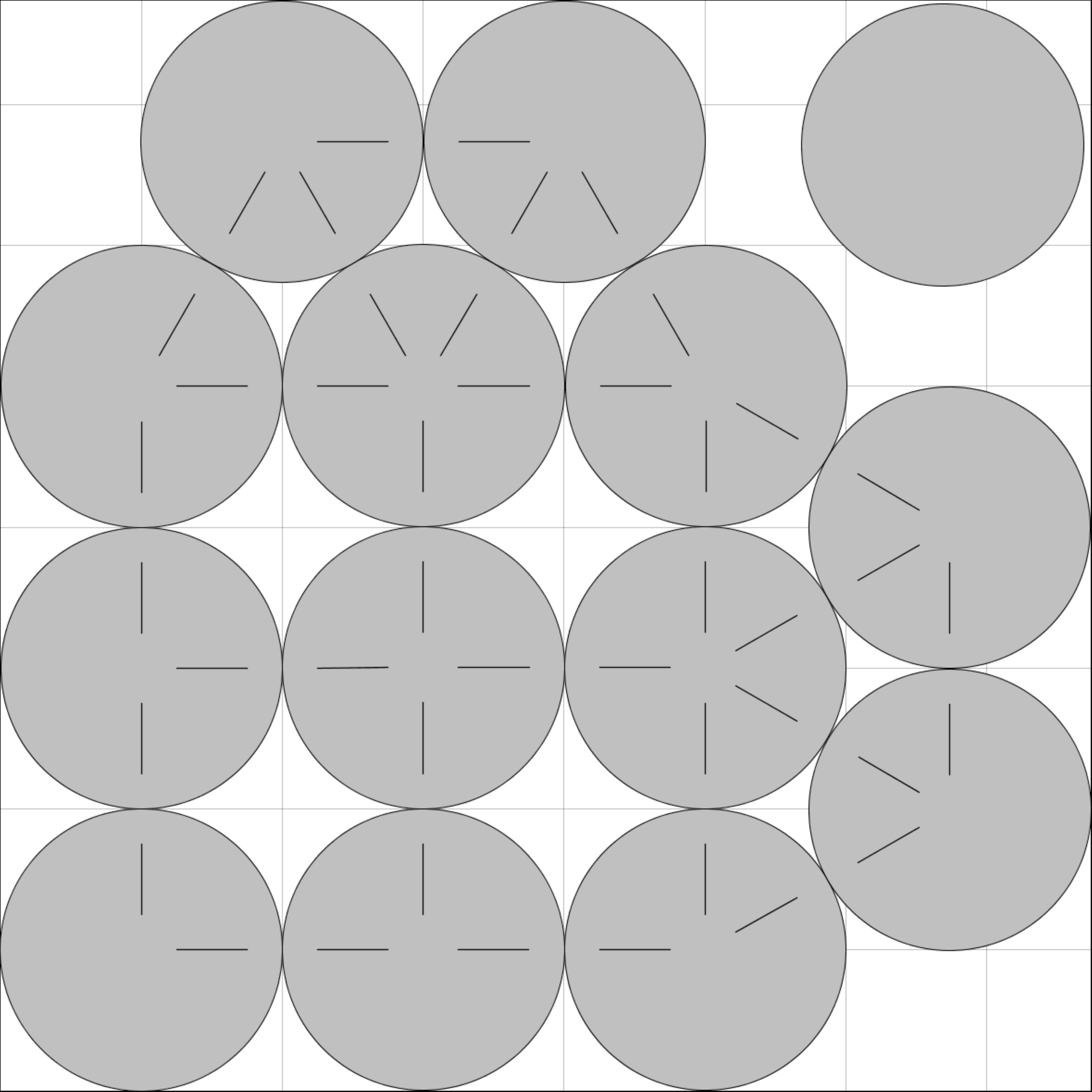}
\end{center}
\caption{An example packing of 14 circles within a square. Contacting circles are indicated by lines.}
\label{fig:packing-example}
\end{figure}

Circle packing is the problem of positioning a given number of congruent circles in such a way that the circles fit fully in a square without overlapping.
A large number of circles makes finding a solution difficult, due in part to the coexistence of many different circle arrangements with similar density.
For example, the packing in Figure \ref{fig:packing-example} can be rotated across either or both axes, and the free circle in the upper-right corner (a ``free circle'' or ``rattle'') can be moved without affecting the density of the configuration.

To represent a circle-packing instance with $N$ objects as an optimization problem we use $\mathcal{O}(N)$ continuous variables and $\mathcal{O}(N^2)$ constraints.
Each object has 2 variables: one representing each of its coordinates (or, more generally, $d$ variables for packing spheres in $d$ dimensions).
For each object we create a single \emph{box-intersection} constraint, which enforces that the object stays within the box.
Furthermore, for each pair of objects, we create a \emph{pairwise-intersection} constraint, which enforces that no two objects overlap.

Circle packing has been extensively studied in the literature \citep[e.g.][]{Szabo2007}.
Of particular relevance, \citet{Gravel2009} showed that the Divide and Concur (DC) algorithm is an effective algorithm for circle packing; however, those results depended upon an ad-hoc process that dynamically weighted variable updates relative to pairwise object distances.
The algorithmic effect, intuitively, is that circles that are far apart do not inform each others' locations and thus iterations-to-convergence improves dramatically if distant circles have little or no effect on each other, especially when scaling to large numbers of circles.
As the next section demonstrates, we achieve a similar performance improvement compared with ADMM/DC using zero-weight messages in our three-weight algorithm, but the approach is simpler and can be applied more generally to a variety of problems where local constraints that are ``inactive'' could otherwise send messages that would slow progress towards convergence.

\subsection{EVALUATION}

We implemented both types of intersection constraints as cost functions on the left.
\emph{Box-intersection} functions send messages, of standard weight ($\rho_0$), for circles sending $n$ messages outside the box to have an $x$ position at the nearest box boundary.
For those circles that are not outside the box, a zero-weight message is sent to stay at the present position ($\overrightarrow{\rho}_{ij}^{t}=0$ and $x_{ij}^{t}=n_{ij}^{t}$).
\emph{Pairwise-intersection} functions are analogous: intersecting circles are sent standard weight messages reflecting updated $x$ positions obtained by moving each circle along the vector connecting them such that equation (\ref{eq:x_updatewithrho}) is satisfied (if both circles send equal weight messages, they are moved an equal amount; if one has a standard weight and one a zero weight, only the circle sending a zero-weight message is moved), while non-intersecting circles are sent zero-weight messages to remain at their present locations ($x_{ij}^t = n_{ij}^t$). 

We first compared iterations-to-convergence between our algorithm and ADMM for a small number, $N=1\dots24$, of unit circles in a square\footnote{We utilized the optimal box side length for each $N$ as listed on \emph{http://www2.stetson.edu/$\sim$efriedma/cirinsqu}.}.
With 4 random conditions per $N$, $\rho_0= 1$, and $\alpha=0.01$\footnote{We found this value to yield the greatest proportion of converged trials for both algorithms after an empirical sweep of $\alpha \in \{0.001, 0.005, 0.01, 0.05, 0.1, 0.2\}$.}, we found that for $N<20$, our algorithm improved performance infrequently: only 42\% of trials had an improvement of $2\times$ or more, and median improvement in iterations was $1.77\times$.
However, for $N\geq20$, our algorithm showed improvements of $2\times$ or more on 90\% of trials and median improvement was more than $116\times$.

\begin{figure}[tb]
\begin{center}
\includegraphics[width=.99\columnwidth]{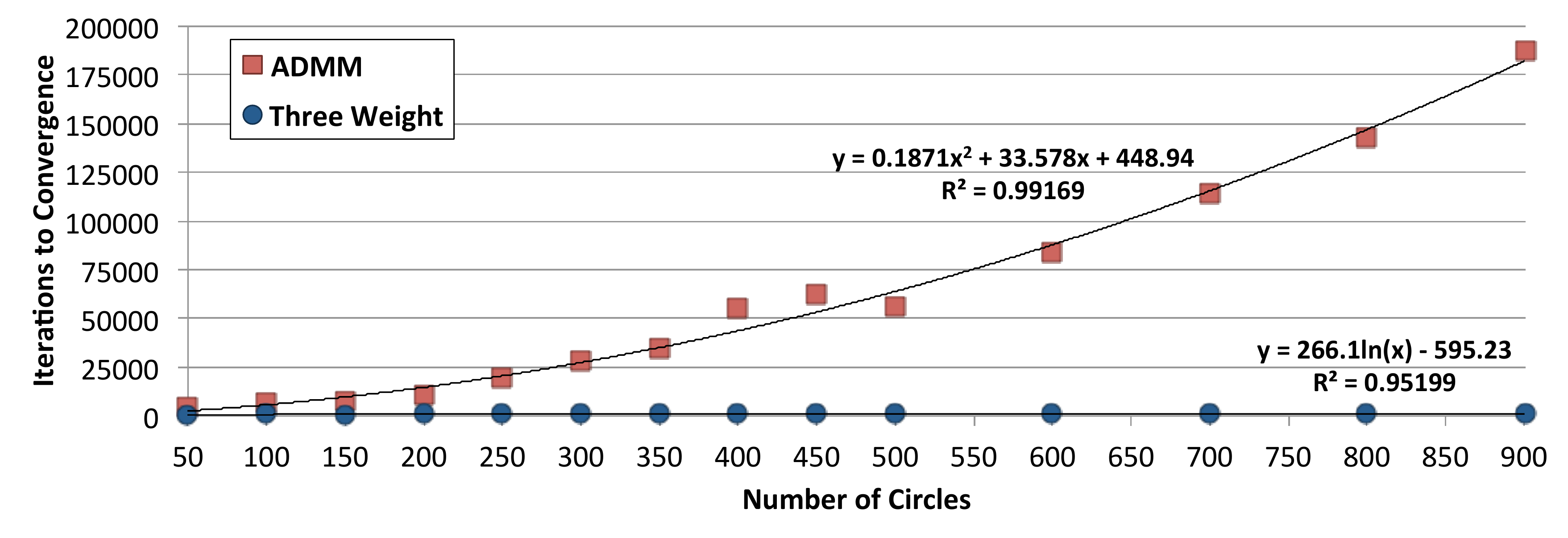}
\end{center}
\caption{Growth in iterations (y-axis) when performing circle packing with large numbers of circles (x-axis) when comparing ADMM with standard weights of 1.0 and our three-weight algorithm.}
\label{fig:packing-big}
\end{figure}

We thus proceeded to compare our algorithm and ADMM as $N$ grew large within a unit square\footnote{To guarantee convergence and reasonable solution time, we took the best known packings from \emph{http://www.packomania.com} and decreased radius by 5\% and used $\alpha=0.01$.}, with results summarized in Figure \ref{fig:packing-big}.
Whereas our algorithm showed only logarithmic growth in iterations as $N$ increased ($R^2>0.95$; likely time required to converge to a desired numerical tolerance of ${10}^{-11}$), and remained fewer than 1220 iterations after $N=901$, ADMM with weights of 1.0 increased quadratically ($R^2>0.99$) and required more than $187,000$ iterations.

We note again that the improvement shown by our three-weight algorithm has a strong intuitive basis. 
In the standard ADMM/DC algorithm, each circle is effectively being sent a message to stay where it is by every other circle in the system that is not overlapping with it, and this can tremendously slow down convergence when $N$ is large. By allowing for zero-weight messages from constraints that do not care about the location of a circle, the algorithm becomes focused on those constraints that actually matter.

\section{COMPARISON WITH BP AND CONCLUSIONS}
\label{sec:compareBP}

We conclude with some general remarks, comparing message-passing algorithms based on ADMM with the family of belief propagation (BP) algorithms.
These families of algorithms are quite similar, in that they can be described in terms of messages passing back and forth between nodes in a factor graph, until (hopefully) convergence is reached, at which point the desired beliefs can be read off. 

The sum-product version of belief propagation, used to compute marginal probabilities in graphical models, has a variational interpretation in that its fixed points correspond to the minimum of the Bethe free energy function \citep{Yedidia2005}.
The ADMM-based message-passing algorithms have an even more straightforward variational interpretation---they are directly minimizing an energy.
It is in fact quite surprising that the DC algorithm, which at first sight seems to be a projection-based constraint satisfaction algorithm, actually can be derived from an energy minimization procedure.

One apparently significant difference is that BP algorithms maintain messages and beliefs that are probability distributions, while the ADMM-based algorithms use messages and beliefs that are normally a single value representing the current best guess for the variable.
However, this difference is not as great as it might appear, especially when indicator variables are used to represent a discrete variable, as in the example of Sudoku.
Thus, the ADMM-based Sudoku algorithms use $N$ binary variables representing the possible state of each cell, so that a collection of $N$ beliefs for one cell have the same dimensionality as a probability distribution that BP would use to represent a belief. 
In fact, when one uses an indicator variable representation, the ADMM-based algorithms use essentially the same memory space as BP algorithms, and really only differ in the update rules.

In our view, the ADMM-based message-passing algorithms have three important advantages over BP algorithms, all exemplified in the circle-packing problem.
The first is that continuous variables are very easy to deal with, in comparison with BP algorithms where quantization of naturally continuous variables must often be used, with a complexity that grows rapidly with the number of quantization levels \citep{Felzenszwalb2006}. BP algorithms exist that deal with continuous variables by sending messages constrained to be Gaussian probability distributions, but the factor graphs that these algorithms can handle only allow for a limited class of possible function costs and constraints \citep{Loeliger2004}.

The second advantage of ADMM-based algorithms is that they easily handle constraints that would be awkward to deal with in BP, such as the constraint that circles cannot overlap. 

The third and perhaps most important advantage is that ADMM-based algorithms will only converge to fixed points that satisfy all the hard constraints in the problem, whereas BP algorithms can converge to fixed points that fail to satisfy all the hard constraints.
This is well known in the case of error-correcting decoders when BP-based decoders can converge to ``pseudo-codewords'' \citep{YedidiaEtAl2011}.
But it is perhaps an even more serious issue in situations, such as circle packing, when BP algorithms converge to {\em non-informative} fixed points.
In particular, a BP-style algorithm for circle packing would begin with messages from the variables representing the circle centers that would effectively say that the circle has an equal chance of being anywhere within the square.
Having received those non-informative messages, the constraints would send out messages that effectively would tell the circles that they have an equal chance of being anywhere within the square, and a non-informative fixed point where all messages continued to give useless flat probability distributions would quickly be reached.
Perhaps a clever scheme could be invented that would avoid this problem, but the general tendency of BP-based algorithms to reach non-informative fixed points in the absence of strong local evidence is a problem that the ADMM-based algorithms gracefully avoid.

To summarize, the ADMM-based message-passing algorithms that we have introduced here, in particular the three-weight version that we have shown potentially solves non-convex problems much faster than Divide and Concur algorithms, have important advantages over the more widely used belief propagation algorithms, and we believe these algorithms have a promising future with many possible applications.

\bibliographystyle{apalike}
\bibliography{ref}

\end{document}